\def\year{2020}\relax
\documentclass[letterpaper]{article} %
\usepackage{aaai20}  %
\usepackage{times}  %
\usepackage{helvet} %
\usepackage{courier}  %
\usepackage[hyphens]{url}  %
\usepackage{graphicx} %
\usepackage{graphicx}  %
\usepackage{xspace}
\usepackage{xcolor}
\usepackage{adjustbox}
\usepackage{array,multirow,graphicx}
\usepackage{capt-of}  %
\usepackage[absolute]{textpos}  %
\usepackage{amsfonts}
\usepackage{siunitx}

\newcommand*{\eg}{\textit{e.g.}\@\xspace}
\newcommand*{\ie}{\textit{i.e.}\@\xspace}

\def\plc/{projection-lifting cycle}
\def\calA/{\mathcal{A}}
\def\calB/{\mathcal{B}}
\def\calC/{\mathcal{C}}

\def\domA/{$\mathcal{I\!\!M}$}
\def\domB/{$\mathcal{P\!\!R}$}
\def\domC/{$\mathcal{T\!\!R}$}

\title{Chained Representation Cycling: Learning to Estimate 3D Human Pose and Shape by Cycling Between Representations}
 
\begin{document}

\twocolumn[{%
  \renewcommand\twocolumn[1][]{#1}%
  \maketitle
\author{\vspace*{1cm}}
\begin{textblock*}{\textwidth} (1.8cm, 4.8cm) {
    \noindent
    \begin{center}
      \begin{tabular}[t]{c}
        Nadine R\"uegg\textsuperscript{1, 2}\\
        {\tt\small nrueegg@tue.mpg.de}
      \end{tabular}
      \quad
      \begin{tabular}[t]{c}
        Christoph Lassner\textsuperscript{3}\\
        {\tt\small classner@tue.mpg.de}
      \end{tabular}
      \quad
      \begin{tabular}[t]{c}
        Michael Black\textsuperscript{2}\\
        {\tt\small black@tue.mpg.de}
      \end{tabular}
      \quad
      \begin{tabular}[t]{c}
        Konrad Schindler\textsuperscript{1}\\
        {\tt\small schindler@ethz.ch}
      \end{tabular}
    \end{center}
  }
\end{textblock*}
\begin{textblock*}{\textwidth} (2cm, 5.9cm) {
    \noindent
    \begin{center}
      \textsuperscript{1}ETH Z\"urich\quad
      \textsuperscript{2}Max Planck Institute for Intelligent Systems, T\"ubingen\quad
      \textsuperscript{3}Amazon
    \end{center}
  }
\end{textblock*}  
  \vspace*{1cm}
  \begin{center}
    \includegraphics[width=0.65\textwidth,clip,trim=0cm 8.5cm 8.3cm 0cm]{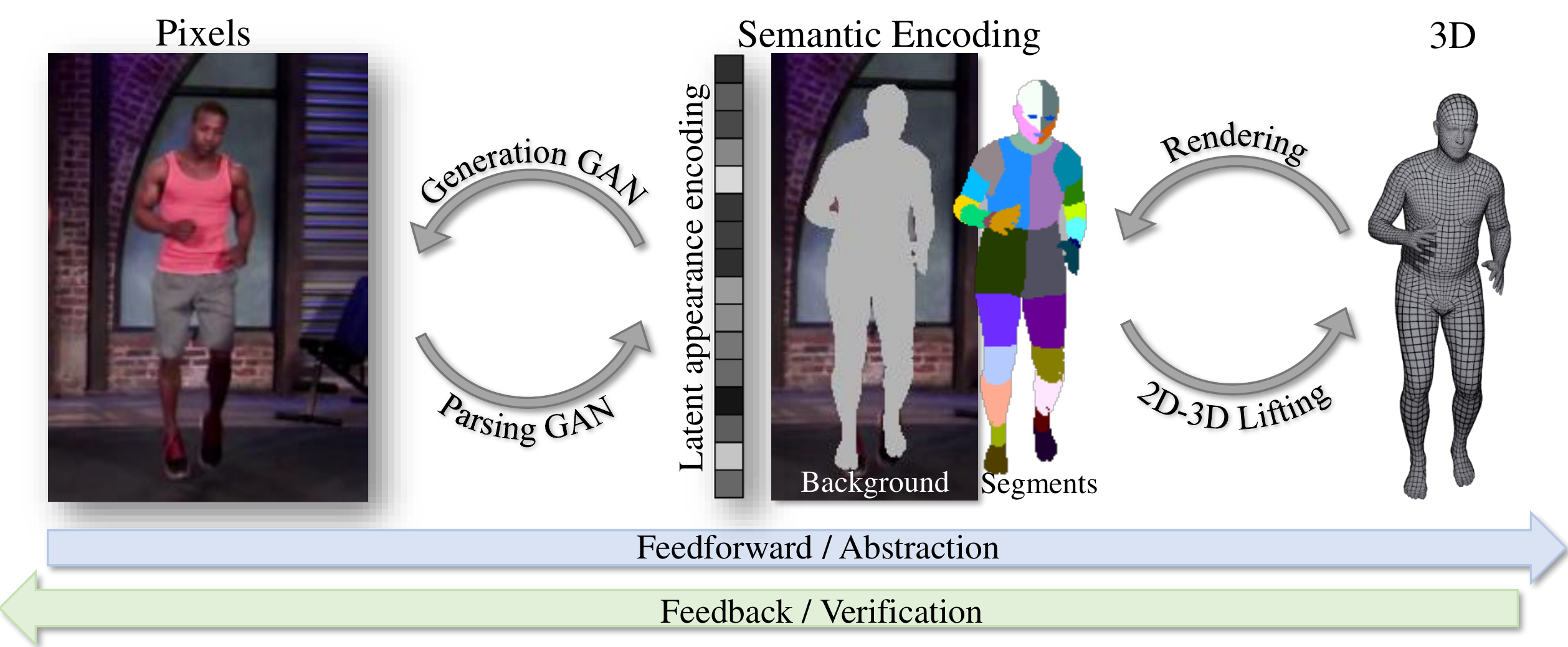}
    \vspace*{-0.6cm}
  \end{center}
  \begin{center}
    \captionof{figure}{\textit{We present a method to learn a mapping between 2D pixels and deformable 3D models in an unsupervised way.} We achieve this by using a chained cycle architecture
between 2D images, a semantic encoding as latent appearance vector, background and segments,  and the parameters of a 3D body model.
  Our model can be used in both directions: to automatically recover 3D parameters from 2D data and to generate renderings with varying appearance.}
    \label{fig:teaser}
\end{center}
}]

\begin{abstract}

The goal of many computer vision systems is to transform image pixels
into 3D representations. Recent popular models use neural networks to
regress directly from pixels to 3D object parameters. Such an approach
works well when supervision is available, but in problems like human
pose and shape estimation, it is difficult to obtain natural images
with 3D ground truth.  To go one step further, we propose a new
architecture that facilitates unsupervised, or lightly supervised,
learning.  The idea is to break the problem into a series of
transformations between increasingly abstract representations.  Each
step involves a cycle designed to be learnable without annotated
training data, and the chain of cycles delivers the final solution.
Specifically, we use 2D body part segments as an intermediate
representation that contains enough information to be lifted to 3D,
and at the same time is simple enough to be learned in an unsupervised
way.  We demonstrate the method by learning 3D human pose and
shape from un-paired and un-annotated images.  We also explore varying
amounts of paired data and show that cycling greatly alleviates the
need for paired data.  While we present results for modeling humans,
our formulation is general and can be applied to other vision
problems.

\end{abstract}

\section{Introduction}

A fundamental task of any vision system, whether natural or
artificial, is to learn the relationship between the observed
``image'' and the 3D scene structure. That is, the system must
transform pixels into 3D information and, conversely, take 3D
hypotheses about the world and confirm them in the image. This can be
thought of as a ``cycle'' from image pixels to 3D models. Learning
such a cyclic mapping is challenging and current methods
assume paired training data of image pixels and ground truth 3D.  We
present an approach that requires little, or no, paired data.

To make the problem concrete, we consider the problem of 3D human pose
and shape estimation. This is a challenging test domain since the
human body is complex, deformable, and articulated. Additionally,
it varies dramatically in appearance due to clothing, hairstyles,
etc.  If we can find a solution to learn 3D body shape and pose from
pixels, then we posit that this approach should be powerful enough to
generalize to many other scenarios.

Consider the example in Fig.~\ref{fig:teaser}. Here, we have three
representations of the human body: pixels, segment-based, semantic 2D
maps, and 3D models.  The goal is to learn the mapping from image
pixels to 3D models, but we take a path through an intermediate
representation and argue that this sequence is easier to learn. The
key question is: what properties should intermediate representations
in the chain have to facilitate learning?  We suggest that a good
intermediate representation should be pixel-based, so that the system
can learn a per-pixel mapping from the image pixels to the
representation.  That is, our part segmentations are not so abstract
that the mapping becomes too hard to learn in an unsupervised way.
Second, the intermediate representation should support 3D inference.
That is, it should carry enough information that it is easy to learn
the mapping to 3D pose and shape.  Previous work has shown that it is
possible to learn the mapping that ``lifts'' human part segmentations
to 3D human pose~\cite{lassner2017unite,omran2018neural}.  We go
further to show that we can learn a mapping from image pixels via the
intermediate representation to 3D, and vice versa, with little or no
training data.

Specifically, imagine you are given a set of images, \domA/, showing people
with arbitrary pose and appearance. Moreover, you have access to a
``high-level''  3D human body model.
Rendering photo-realistic training images from a 3D body remains a
hard problem, while it is easy to render the body parts as a segmentation mask (see Figure \ref{fig:teaser}).
But we do not know the relationship between the set of rendered
segmentations, \domB/, and the natural image set \domA/.
Can we nevertheless use the two unrelated sets of pictures to learn
the mapping from raw pixels in \domA/ to 3D body shape and
pose? To do this we develop a novel, unsupervised multi-step cycle model---a representation chain---between three different representations: pixels, 
segments, and 3D parameters.

The presented approach provides a step towards unsupervised learning
of 3D object pose from images.  A key advantage of such a direction is
that it theoretically allows us to train from unlimited amounts of
data, which should improve robustness to real imaging conditions and
natural variability.

To test the effectiveness of the approach, we experiment on human pose
estimation using both synthetic and real datasets.  While not yet as
accurate as fully supervised methods, we show that it is feasible to
learn with no paired image and 3D data.  To our knowledge, this is the
first approach to tackle this problem for highly deformable 3D models
with such large variability in appearance.  We also evaluate the
impact of small (\ie, practical) amounts of supervision, which
improves performance.  Additionally, we show how the model learns a
representation that can be used for other tasks like sampling
synthetic bodies, transferring clothing, and reposing.
Our hope is that the framework of \emph{chained representation
  cycling} inspires other applications and research into new
intermediate representations.

\section{Related Work}

\subsubsection{Domain adaptation with CycleGAN.}
A number of recent works employ cyclic adversarial learning to
translate between different image domains~\cite{hoffman2017cycada,CycleGAN2017,lee2018diverse,hoshen2018nam,Mueller_2018_CVPR}.
\emph{CyCADA}~\cite{hoffman2017cycada}, the most relevant representative of that line of work for our
task, is trained on pairs of synthetic
street scenes and rendered images as well
as unpaired real images.%
\cite{zhu2017toward} introduce \emph{BicycleGAN}, a method for
cross-domain image-to-image translation that includes a latent vector
to encode appearance information, as in our work. In contrast to ours,
\emph{BicycleGAN} needs supervision by paired data from the two
domains.
\emph{Augmented
  CycleGAN}~\cite{almahairi2018augmented} and
\emph{MUNIT}~\cite{huang2018multimodal} also advocate the use of
latent vectors to separate appearance from content.
We have tested the \emph{MUNIT} architecture for our problem, but found
that it did not perform better than a vanilla \emph{CycleGAN}.

\subsubsection{3D Human Pose and Shape Estimation.}
A full review of 3D pose and shape estimation models is beyond the scope of this paper. Recent, representative examples
include~\cite{sun2017compositional,tome2017lifting,pavlakos2017volumetric,dmhs_cvpr17,pavlakos2018ordinal,kanazawa2018end,omran2018neural}.
Typically, the output is a set of key joints linked corresponding to a stick figure. These reconstructions often
result in unrealistic 3D bone lengths.
One solution is to fit a complete 3D mesh instead of just a skeleton. This makes it possible to exploit
silhouettes to better constrain the part localization in image
space~\cite{lassner2017unite}, and can also include a prior that
favors plausible shapes and poses~\cite{Bogo:ECCV:2016}, but requires a supervised method to estimate key locations.

Even when using an elaborate body model, the intermediate 2D
representation of the body is predominantly a set of
keypoints~\cite{tome2017lifting,martinez2017simple,kanazawa2018end}. We deviate from that tradition and instead model
the body as a collection of 2D segments corresponding to body parts,
like in~\cite{omran2018neural}.

Another method in the context of unsupervised 
learning is~\cite{balakrishnan2018synthesizing}, which
performs direct synthesis without explicit 3D reconstruction.
A different, more expressive approach is to model
directly the volumetric body shape, but also deviate
from true human shapes~\cite{varol18_bodynet}.

\subsubsection{Generative Person Models.}
One of the first end-to-end trainable generative models of people was
presented in~\cite{Lassner:GP:2017}. \cite{ma2017pose}~and
\cite{Esser_2018_CVPR}~focus on transferring person appearance to a
new pose, which we also showcase with our model.
\cite{ma2018disentangled}~aims for a factorization of foreground,
background and pose, similar to our approach, but only tackles the
generative part. In contrast to the aforementioned models,
\cite{Zanfir_2018_CVPR}~takes a high-fidelity approach to human
appearance transfer, combining deep learning with computer graphics
models. To our best knowledge, all
existing methods for appearance transfer rely on a keypoint detector
or part segments, and therefore require either labeled data or
multiple images of the same person for training.

\subsubsection{Semi-Supervised Approaches.}
A major difficulty for 3D body fitting is to obtain enough ground
truth 3D poses. Several recent works therefore
aim to do the 2D-to-3D lifting with little or no 3D supervision.
\cite{kanazawa2018end} cast the lifting problem as 3D pose regression and
train it by minimizing a reprojection loss in 2D and an adversarial
loss on the predicted 3D pose.
Related is~\cite{kudo2018unsupervised}, which follows the same
general idea, but reprojects the predicted 3D joints into a number of
2D views and places the adversarial loss on the resulting 2D joint
configurations.
\cite{rhodin2018unsupervised}~train an encoder-decoder architecture to
synthesize novel viewpoints of a person. Their pretraining only needs
multi-view data, but no annotations. \cite{mvcTulsiani18}~follows the same idea and uses solely
multiview consistency as a supervisory signal, for rigid objects.

Recently, \cite{lorenz2019unsupervised} have presented a method for
unsupervised, part-based disentangeling of object shape and appearance,
given video data or images with a plain white
background. \cite{jakab2018unsupervised,jakab2019learning,minderer2019unsupervised}
also propose strategies for unsupervised detection of 2D human keypoints in videos.
None of those approaches treats the background explicitly, so training is only
possible on videos with almost no camera motion.

\section{Foundations}

In this section we briefly introduce two prior methods used as components of our models.

\subsubsection{Skinned Multi-Person Linear Model (SMPL).}  %
Our method learns a mapping between the parameterization space of a 3D model and 2D images
of object instances of the model. As 3D model in our work, we choose SMPL, the ``Skinned
Multi-Person Linear Model'' as proposed in ~\cite{loper2015smpl}. This model establishes
a pose and shape space for human bodies in 3D space. Concretely, it is parameterized by
the pose parameters $\theta \in\mathbb{R}^{72}$ (one 3D axis angle rotation vector for 23 body
parts plus a root rotation vector) and the shape parameters $\beta \in\mathbb{R}^{10}$ (ten
PCA components to account for variation in body shape). The result of posing and shaping SMPL
is a \num{6890} vertex mesh with \num{10} faces.

\subsubsection{Neural Body Fitting (NBF).}
NBF describes an approach for supervised 3D human pose estimation \cite{omran2018neural}. It uses (different) body part segments as an intermediate representation and maps these segments to a parameterization of SMPL.
This mapping is performed using a neural network that takes a $224\times224\times3$ color
coded segmentation map as input and predicts the full SMPL parameters. This module can be treated
as a black box for our approach, and we use it to implement the translation between intermediate representation and 3D space, with changes only to input and output encoding and losses.

\section{Approach}
\label{sec:approach}

We establish a connection between the 2D and 3D representations of
objects, and present a novel, unsupervised approach for mapping 3D
objects to images and back.
We do note that for many object types 3D models exist, or can be created from existing 3D data in an automated manner. This includes objects with limited shape variability and articulation (\eg, cars, furniture), but more recently also common animals \cite{zuffi20173d}. 
As we aim to handle articulated pose, including 3D orientation and
translation, the projected images can vary strongly in appearance. To
still be able to use unsupervised training, we need to reduce the
complexity of the task as much as possible.
An ideal intermediate representation would describe the image
and, at the same time, reduce the complexity of the following,
chained step. We introduce part segments as this main, intermediate
representation to connect the two domains. With them, we can
``factorize'' an image into background, a segment-based description of
object pose and shape, and residual appearance. With this
representation, (1) it is trivial to extract the background; (2)
object pose and shape are encoded more richly than by
keypoints~\cite{omran2018neural} for the next step; and (3) residual
appearance reduces to precisely the information needed to ``paint'' a
person with known segmentation. Furthermore, the part segments reside
in the same domain as images: the 2D plane. They are spatially related
to the image pixels and are therefore well-predictable by a
CNN.

\subsection{Domains}
\label{sec:data-types}

For clarity we refer to the 2D image space as domain~\domA/~(image), to the
intermediate representation as domain~\domB/~(proxy), and to the observed 3D
world as domain~\domC/~(target).
Domain~\domA/ consists of images of the object class of interest, here
people. Domain~\domC/ is represented with a fully parameterized,
deformable 3D model of that class. For our scenario we use the SMPL
body model~\cite{loper2015smpl} with pose and body shape
parameters.
Domain \domB/ factors the image space into a \emph{background},
explicit \emph{part segments} and a \emph{latent appearance
  code}. Technically, we store this information as (1) a 4-channel
image that contains in the first three channels the natural image of the background region overlayed by a colour-coded segment map of the foreground (with arbitrary, well-separated colours) as well as in channel four a binary foreground-background mask and (2) a vector
$z$ that encodes appearance.

\subsection{Cycle \domA/-\domB/}
\label{sec:cycle}

The cycle \domA/-\domB/ consists of a \textit{parsing GAN} model, translating images to part segments, background and appearance, and a \textit{generative GAN} model to translate this intermediate representation to images. We base the design of this loop on the CycleGAN architecture~\cite{CycleGAN2017}. 

\subsubsection{CycleGANs.}
\label{sec:cyclegan}
are a model family designed such that a mapping from one image domain~$\calA/$ to another image domain~$\calB/$ can be found in an unsupervised way. The mappings $G : \calA/ \rightarrow \calB/$ and $F : \calB/ \rightarrow \calA/$ are learned using two types of loss functions:
\emph{(i)} Adversarial losses encourage images from $G(\calA/)$ to be indistinguishable from images sampled from domain~$\calB/$. Similarly, an adversarial loss enforces matching distributions for $F(\calB/)$ and domain~$\calA/$.
\emph{(ii)} Cycle consistency losses enforce $F(G(\calA/)) \approx \calA/$ and $G(F(\calB/)) \approx \calB/$.
Through the combination of the two losses, it is possible to sidestep annotations altogether: the adversarial losses ensure that ``translated'' representations match the target domain, and the consistency losses enforce a consistent mapping between domains. In the following, we extend this basic CycleGAN model for our purpose, to enforce the intended factorization of images from \domA/ in \domB/.

\subsubsection{A Complete Segmentation Cycle.}
\label{sec:segm-cycle}
We develop a cycle between real-world images and the combination of background, part
segments and residual appearance. In contrast to traditional image
translation, we need multiple components in \domB/ to capture all
relevant content of natural images in \domA/. Only then we can
establish a full translation cycle without loss of information.

\noindent\textbf{Generator from \domA/ to \domB/.}\label{sec:generator-from-a-to-b}
\begin{figure}
  \begin{center}
    \includegraphics[width=0.9\linewidth]{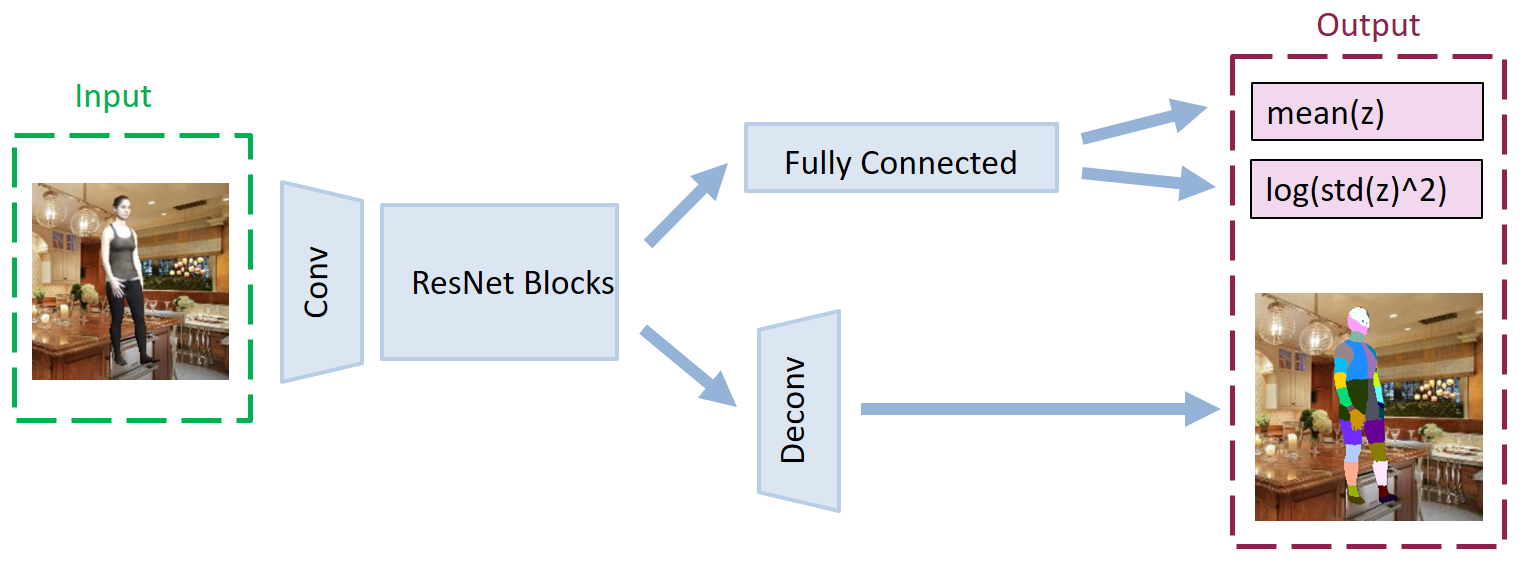}
  \end{center}
    \vspace*{-0.5cm}
  \caption{\textit{The mapping function from \domA/ to \domB/.} We use a classical encoding structure, followed by (1) a fully connected layer predicting mean and variance of the residual appearance as well as (2) a set of deconvolutions to produce a part segmentation map in the original resolution.}
  \label{fig:gen_to_Bz}
\end{figure}
The structure of the generator that maps an image from domain~\domA/
to domain~\domB/ is shown in Fig.~\ref{fig:gen_to_Bz}. The image is
encoded by convolutional layers and processed by a
ResNet~\cite{he2016deep}. The output is processed by several
fractionally strided convolutions to produce a segmentation and background
map at input resolution. In the first three channels, this map contains the color coded semantic segments of the person where applicable and the background colors where not. In the fourth channel, it contains a semantic segmentation with solely foreground and background labels.
In a second branch, we use a fully connected
layer to predict the mean and variance of the a latent person appearance encoding vector. This
parameterization is inspired by variational
autoencoders~\cite{kingma2013auto} and enables us to sample person
appearance vectors for the cycle part \domB/-\domA/ at test time. This allows us to
fully sample representation for \domB/ randomly.

\noindent\textbf{Generator from \domB/ to \domA/.}\label{sec:gen_from_Bz_to_A}
\begin{figure}
  \begin{center}
    \includegraphics[width=0.98\linewidth]{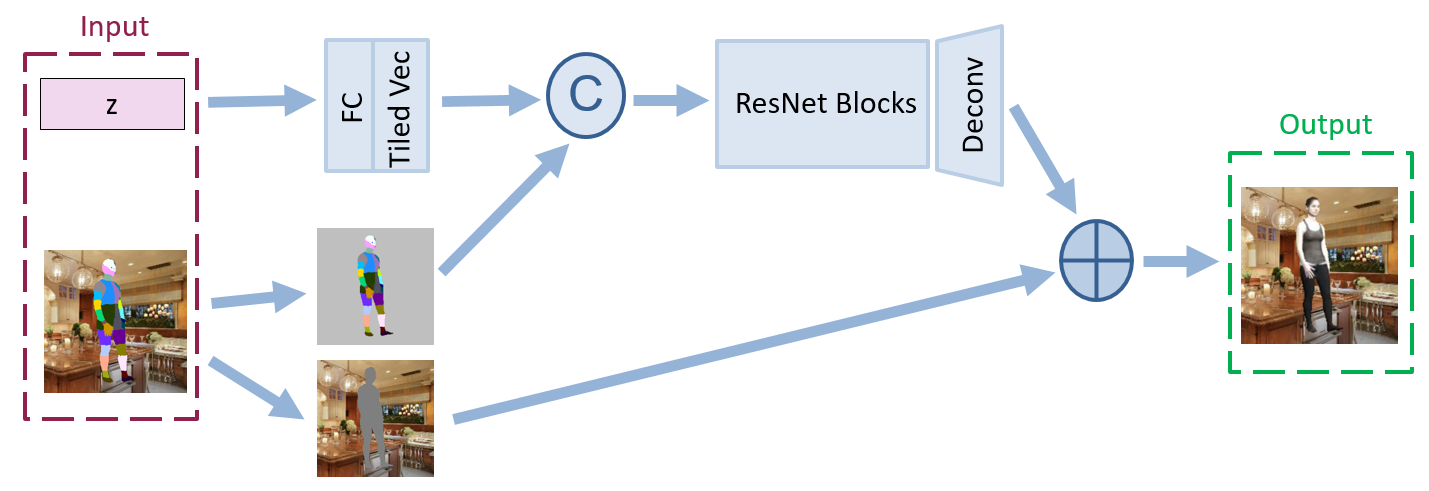}
  \end{center}
    \vspace*{-0.5cm}
  \caption{\textit{The mapping function from domain~\domB/ to domain~\domA/.} The part segment image and the background are split and the part segments are used together with the residual appearance encoding to produce the object image before fusing it with the background.}
  \label{fig:gen_from_Bz}
\end{figure}
The cycle part to generate an image from its semantic description
receives as input a background image, the part segment image and a
vector $z$. 
The segments together with the $z$ vector contain all
information needed to generate an image of a human in the desired pose
and appearance.
As illustrated in Fig.~\ref{fig:gen_from_Bz}, $z$ is tiled and
concatenated with a resized version of the part segmentation map. We
use a ResNet~\cite{he2016deep} followed by fractionally strided
convolutions to produce an image in the original size. The fusion with
the background is guided by the mask from domain~\domB/. %

All design choices for building the generators are driven by two needs: to
be able to randomly sample instances in both domains and to create a representation
in domain \domB/ that is helpful for the following chain link to domain~\domC/, but still
`complete' in the sense that it contains enough information to generate an example
for domain~\domA/.

\noindent\textbf{Mitigating Information Hiding.}\label{sec:mitig-inform-hiding}
We find that, when translating across domains with strongly differing appearance,
models tend to ``hide'' texture information by reducing
its contrast. This happens in subtle ways even when translating
between natural image domains.

Whereas information hiding in regular CycleGANs is mainly a cosmetic
problem, it is fundamentally problematic in our setting: the network
can hide appearance in the part segments instead of using the vector
$z$ to encode it. This (1)~leads to polluted silhouettes with noisy
segment information and (2)~the network becomes agnostic to changes in
$z$. The ability to sample different appearance cues is important,
because it enables us to generate different images for the cycle
\domB/-\domA/-\domB/-(\domC/). Hence, we take several steps to prevent
the network from hiding appearance in the segmentations: first, we
suppress residual color variations within each predicted part by
flooding it with the assigned part color. Second, in the cycle
\domA/-\domB/-\domA/, both, the originally predicted image and the
thresholded and inpainted image are passed to the transformation
\domB/-\domA/.

\subsection{The Leap to 3D}\label{sec:3d-fitting}

The second part of the chained model is trained to bridge the domains
\domB/ and \domC/, \ie, to match 3D object parameters and part
segment maps. Whereas the link \domC/-\domB/ is implemented as
rendering of a segment color-coded SMPL mesh, we implement the link \domB/-\domC/ as an adaptation of neural body fitting
(NBF~\cite{omran2018neural}), with the part segments in front of
a neutral background as input. Global translation, rotation and shape
parameters are estimated in addition to one rotation matrix per
joint. In order to handle the ambiguity between model size and
camera viewpoint, we height-normalize our 3D model
templates and use depth as the only way to influence the size
of the depicted model.

The \domB/-\domC/ part of our model is first trained on sampled pairs of
3D parameters and rendered part segment maps.
However, the sampled part segmentation maps  in \domB/ are noise-free, as opposed
to the ones produced by our encoder \domA/-\domB/.
Still, we can train our full network by using
the chain (\domC/-)\domB/-\domA/-\domB/-\domC/
in an unsupervised way and enforce a
``reconstruction loss'' for the 3D model parameters.

For this extended cycle \domB/-\domA/-\domB/-\domC/, we use all loss
terms for the \domA/-\domB/ cycle and an additional term for the
reconstruction of 3D parameters. The initial $z$ for the transition
\domB/-\domA/ is \emph{not} sampled randomly, but copied from a
different input image picked at random, to ensure as realistic person
appearance as possible. Furthermore, we stop the 3D reconstruction loss gradients at
domain~\domA/, such that the network is not encouraged to paint
silhouette information into images from  domain \domA/.

\section{Experiments}\label{sec:experiments}

\begin{figure*}
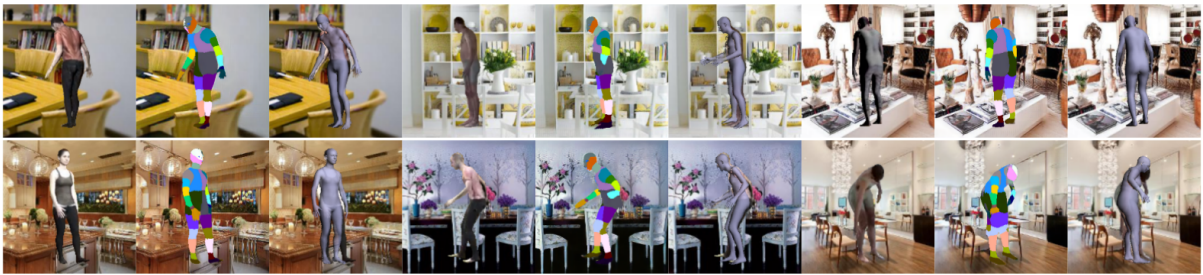

  \begin{center}
\resizebox{0.9\textwidth}{!}{\includegraphics[height=3.5cm]{{{199_results_surreal}}}}
\end{center}
\vspace*{-0.5cm}
\caption{\textit{Qualitative results of our unsupervised model on the
    SURREAL dataset~~\cite{varol2017learning}.} \textbf{Per example, per
  column}: input image, predicted segments, fitted 3D body. Even though
  many body fits are correct w.r.t. the segmentation, the model is susceptible to left-right swaps. The overall body shape is typically
  well-explained.}
\label{fig:surreal_quantitative}
\end{figure*}

In the following, we first present a detailed evaluation on the
synthetic \textit{SURREAL}~\cite{varol2017learning} dataset and
discuss in a second part results on images of real people. Those
images come from the \textit{Human3.6M}~\cite{h36m_pami} and
\textit{Penn Action}~\cite{zhang2013actemes} datasets.

\subsection{Results on the SURREAL Dataset}\label{sec:results}

SURREAL contains images of people rendered with 772
clothing textures and 193 CAESAR dataset textures~\cite{robinette1999caesar}.
Corresponding annotations, including pose, depth maps and segmentation
masks are provided.
To create a realistic scenario,
we process the training dataset such that we have the two unpaired
sets of images, strictly separated by motion sequence+rotation. From
one set we draw images for domain~\domA/, from the other poses for
domain~\domB/.
First, our method is evaluated in terms of semantic body part segmentation and 3D pose error. Second, we provide qualitative fitting results and show examples for appearance transfer and body shape estimation.

\subsubsection{Quantitative Results.}
To our knowledge there are few different papers that present pose estimation results
on SURREAL, all using full supervision \cite{NIPS2017_7108,varol2017learning,varol18_bodynet,madadi2018smplr}. As a connecting
baseline we run a fully supervised training of our model. Note,
though, as our version of the dataset is split to be unpaired, we use
only half as much data for training. Moreover, the chained
architecture is not ideal for the fully supervised setting, due to
less direct gradient propagation.

We first evaluate the semantic segmentation results of our model,
which correspond to the results from the cycle part
\domA/-\domB/. Tab.~\ref{tab:res_related_work_surreal} shows results
for segmentation into 14 parts as well as 3D pose
estimation. \cite{madadi2018smplr} and \cite{varol18_bodynet} list the
mean per joint position error (MPJPE) in millimeter (mm) after the
positions are subtracted from the root joint. We additionally provide the
translation normalized root mean squared error (t-RMSE).
Even though we use a smaller training set and the chained model is not designed for supervised learning, its segmentation results are on par with \cite{varol2017learning} and
\cite{varol18_bodynet}, pose and shape reconstructions are slightly
worse.

\begin{table}
  \begin{center}
    \resizebox{0.48\textwidth}{!}{
      \begin{tabular}{|l|c|c|c|}
\cline{2-4} 
\multicolumn{1}{c|}{} & Seg. (IoU) & t-RMSE &  MPJPE \\
\hline 
SURREAL \cite{varol2017learning} & 0.69 & - & - \\
\hline 
BodyNet \cite{varol18_bodynet} & 0.69 & - & 40.8 \\
\hline 
SMPLR \cite{madadi2018smplr} & - & - & 55.8 \\
\hline 
Ours (fully supervised) & 0.69 & 85.0 & 69.5 \\
\hline 
Ours (unsupervised) & 0.36 & 204.0 & 167.3 \\
\hline 
      \end{tabular}
    }
  \end{center}
  \vspace*{-0.3cm}
  \caption{\textit{Quantitative comparison with supervised methods.}
  We only use half of the dataset for supervised training and still manage to match
  the performance of SURREAL and BodyNet models in segmentation.
  }
  \label{tab:res_related_work_surreal}
\end{table}

\newcommand{\STAB}[1]{\begin{tabular}{@{}c@{}}#1\end{tabular}}
  \renewcommand{\arraystretch}{1.2}
  \begin{table}[!h]
    \centering
     \resizebox{0.48\textwidth}{!}{
    \begin{tabular}{|c|c|r|r|r|r|r|}
      \cline{3-7}
      \multicolumn{2}{c|}{} & \multicolumn{1}{c|}{no sup.} & \multicolumn{4}{c|}{supervision} \\ \cline{3-7}
       \multicolumn{2}{c|}{} & \multicolumn{1}{c|}{-} & \multicolumn{1}{c|}{0.2\%} & \multicolumn{1}{c|}{0.5\%} & \multicolumn{1}{c|}{1\%} & \multicolumn{1}{c|}{100\%} \\ \hline
      
      \multirow{3}{*}{\rotatebox[origin=c]{0}{\parbox[c]{1cm}{\centering   \footnotesize{segmen-tation}   }}}  
      & \STAB{\rotatebox[origin=c]{0}{  14seg  }}  			& 0.364 	& 0.488   & 0.563   & 0.622   & 0.688                          \\
      & \STAB{\rotatebox[origin=c]{0}{  4seg  }}   			& 0.713 	& 0.791   & 0.755   & 0.777   & 0.779                          \\
      & \STAB{\rotatebox[origin=c]{0}{  1seg  }}   			& 0.887 	& 0.924   & 0.923   & 0.932   & 0.943                          \\ \hline
    
      \multirow{3}{*}{\rotatebox[origin=c]{0}{\parbox[c]{1cm}{\centering  \mbox{\footnotesize{3D error}} \mbox{\footnotesize{(normal)}}  }}}  
      & \STAB{\rotatebox[origin=c]{0}{  \small{RMSE}  }}     & 252.90     & 182.95   & 161.74   & 149.72   & 129.96                          \\
      & \STAB{\rotatebox[origin=c]{0}{  \small{t-RMSE}  }}   & 204.29     & 141.90   & 113.79   & 100.71   & 84.97                          \\
      & \STAB{\rotatebox[origin=c]{0}{  \small{tr-RMSE}  }}  & 135.59     & 104.18   & 93.80    & 84.31    & 71.90                           \\ \hline
    
      \multirow{3}{*}{\rotatebox[origin=c]{0}{\parbox[c]{1cm}{\centering  \mbox{\footnotesize{3D error}} \mbox{\footnotesize{(best of}} \mbox{\footnotesize{4 swaps)}}  }}} 
      & \STAB{\rotatebox[origin=c]{0}{  \small{ RMSE}  }}    & 175.81     & 145.88   & 142.45   & 135.47   & 121.75                          \\
      & \STAB{\rotatebox[origin=c]{0}{  \small{t-RMSE}  }}   & 125.88     & 104.28   & 99.69    & 90.84    & 81.51                          \\
      & \STAB{\rotatebox[origin=c]{0}{  \small{tr-RMSE}  }}  & 103.18     & 86.00    & 83.89    & 77.03    & 69.29                           \\ \hline
     
    \end{tabular}
    }
    \vspace*{-0.3cm}
     \caption{\textit{Unsupervised and semi-supervised evaluation
         results.} Ablation study for varying amounts of supervision
       (0\%, 0.2\%, 0.5\%, 1\% and 100\%).
     }
    \label{tab:res_part_supervision}
  \end{table} 
  
\noindent\textbf{Unsupervised and Semi-Supervised Training.}
We now turn to the unsupervised setting. As an ablation study, we include
a semi-supervised variant with small amounts of supervision added.
The results can be found in Tab.~\ref{tab:res_part_supervision}.
We used a fully unsupervised trained model for initialization and then ran training using an L2 loss for the part segment maps of every $k$-th example ($k\in \{500, 200, 100\}$). The 1\% data split contains the 0.5\% as well as 0.2\% data splits completely.
For evaluation, we used the
metrics:

\noindent \emph{Segmentation.} We provide intersection over union
(IoU) scores for the predicted segmentation masks, ignoring the
dominant background category. We additionally provide segmentation
results after combining segments, resulting in a total of 14, 4 or 1
segment. \textit{4seg} corresponds to the segments $d=\{arms, legs,
upper\_body, head\}$, while \textit{1seg} corresponds to
foreground-background segmentation.

\noindent \emph{3D Error (normal).} We report the root mean square
error (RMSE) between predicted and ground truth 3D SMPL joint
locations. Furthermore, we calculate the RMSE with corrected
translation (t-RMSE) and corrected translation as well as rotation
(tr-RMSE). tr-RMSE is equal to the pose error calculated after a Procrustes
matching.

\noindent \emph{3D Error (best of 4 swaps).} To
analyze the effect of the well-known left-right swapping ambiguity, we
provide this additional score. To calculate it, we swap the left and
right arms, respectively legs, in the segmentation; fit the 3D model
for each of the four possible combinations; and use the best fit.

\noindent\textbf{Discussion.} We find that the fully unsupervised learning of segments as well as 3D fits
works surprisingly well, but is often subject to left-right
ambiguities. A main source of information for the model in the
unsupervised case is the outer silhouette of the person. Since
left-right swaps preserve the outer silhouette, it is hard for the model to identify
these swapping errors. This becomes obvious when comparing, \eg,
t-RMSE (best of 4) unsupervised with the t-RMSE (normal) with 100\%
supervision: the unsupervised model learns to estimate body pose and
shape nearly as as well as in the supervised case, but the left-right
swaps remain a hurdle.
Importantly, we find that adding very little supervision can
give just the right hints to overcome this hurdle. When adding even
just 1\% of supervision, the t-RMSE (normal) is halved. Also, with
1\% supervision the segmentation (14 segments) reaches 90\% of the
fully supervised performance.

\begin{figure}
  \begin{center}
\resizebox{0.45\textwidth}{!}{\includegraphics[height=3.5cm]{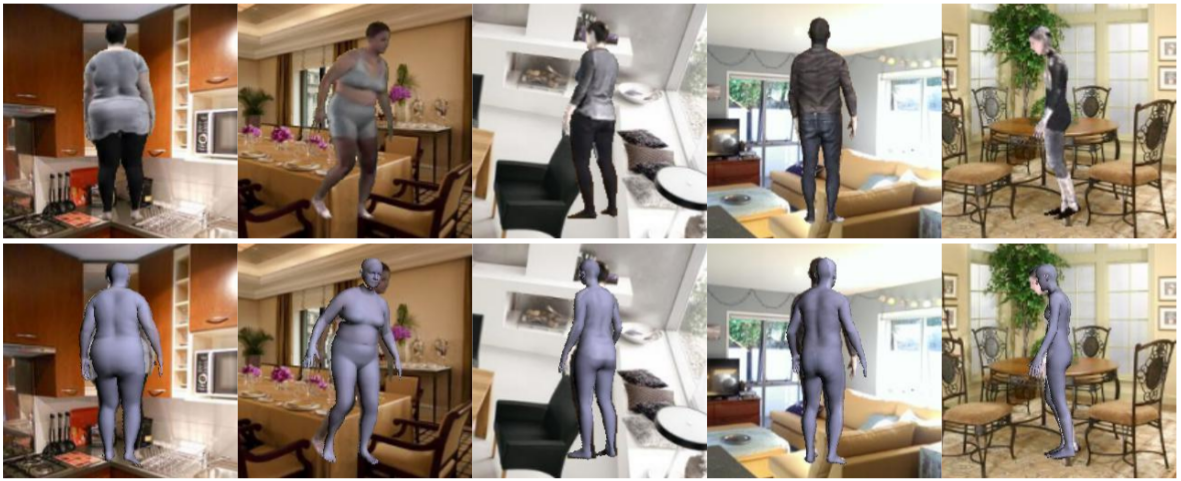}}
  \end{center}

\vspace*{-0.5cm}
\caption{\textit{3D pose and shape predictions of our unsupervised model for people with strongly varying body shape.}}
\label{fig:qualitative_shape}
\end{figure}

\subsubsection{Qualitative Results.}\label{sec:results_surreal_qual}
Our representation bridges all three domains \domA/, \domB/ and \domC/. We
show examples for different modes of operation,
(1)~3D fits from images (the direction \domA/-\domC/);
and (2)~appearance transfer (the direction \domC/-\domA/). All results provided in this section are obtained with
unsupervised training on the SURREAL dataset.

\noindent\textbf{From \domA/ to \domC/ -- 3D Fits and Segments.}
Randomly selected results of the unsupervised, chained
cycle model are shown in Fig.~\ref{fig:surreal_quantitative}. The
explanation of the human silhouette is usually
successful, however the model is susceptible to left-right swaps that have a strong influence on the 3D fit, as discussed above.
Fig.~\ref{fig:qualitative_shape} shows 3D predictions for people with
varying body shape. The silhouette provides a strong
signal to estimate body shape from all directions. This enables the
model to predict shape parameters faithfully, even without any
supervision.

\begin{figure}
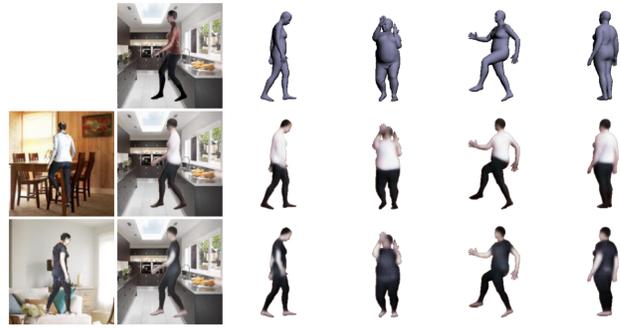

  \begin{center}
  \resizebox{0.48\textwidth}{!}{\includegraphics[height=3.0cm]{{{199_surreal_textures}}}}
\end{center}
\vspace*{-0.5cm}
\caption{\textit{Example results for appearance transfer.}
  \textbf{1$^{st}$ column:} input images. \textbf{2$^{nd}$
    column:} appearance transfer to a new image. The new image is
  visible in the top row, the altered appearance
  below. \textbf{Remaining columns:} the top row shows plain SMPL
  instance, below are results of mapping \domC/-\domA/ with the extracted
  appearance.}
\label{fig:texture_transfer_small}
\end{figure}

\noindent\textbf{From \domC/ to \domA/ -- Appearance Transfer.}
We showcase a different mode of operation for our model and transfer
the appearance between people in
Fig.~\ref{fig:texture_transfer_small}. Appearance $z$ is predicted for
each of the input images on the left. It is transferred to
another image as well as multiple random 3D configurations.
The results are stable for a fixed vector $z$, which indicates
that the model learns to encode appearance and does not simply
hide cues in the part segmentation map. We are able to
transfer appearance to any 3D configuration and obtain good
results, for varying poses and body shapes. The ability to generate
people in previously unseen poses is important for the
cycle \domA/-\domB/ to work.

\subsection{Results on Real Images}

\begin{figure*}
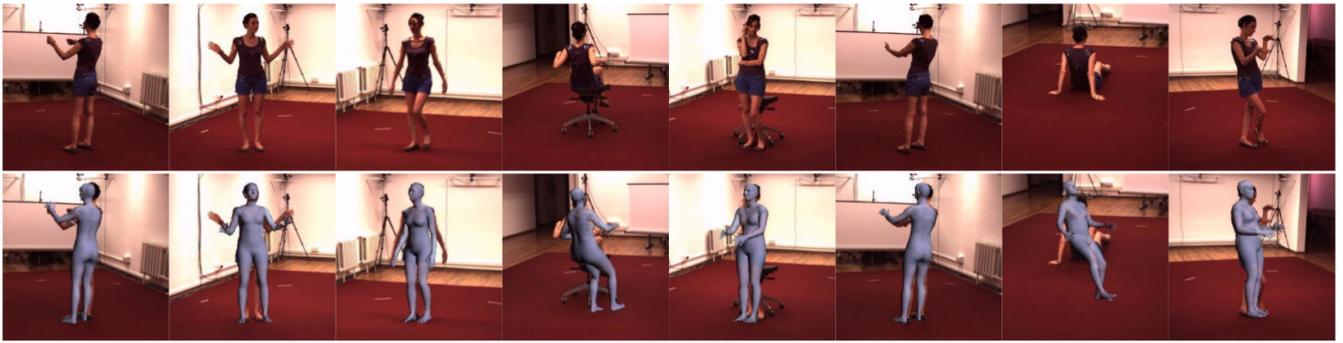

  \begin{center}
\resizebox{\textwidth}{!}{\includegraphics[height=3.5cm]{{{199_results_h36m}}}}
\end{center}
\vspace*{-0.3cm}
\caption{\textit{Random examples of estimated 3D pose and shape on the H36M dataset~\cite{h36m_pami}.} These results are generated by a model that uses partial supervision with only 66 labeled images.}
\label{fig:h36m}
\vspace*{0.3cm}
\end{figure*}

\begin{figure*}
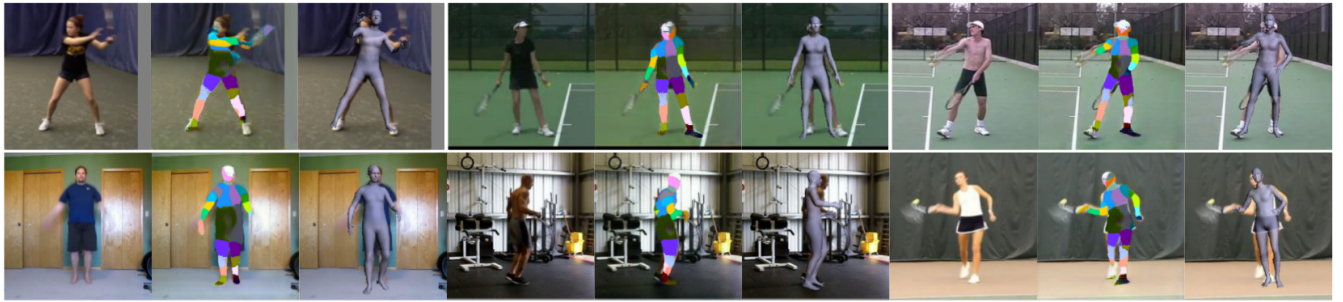

  \begin{center}
  \resizebox{\textwidth}{!}{\includegraphics[height=3.5cm]{{{199_results_pennaction}}}}
  \end{center}
\vspace*{-0.3cm}
\caption{\textit{Qualitative results of our unsupervised model on in-the-wild images \cite{zhang2013actemes}}. \textbf{Per example, per column:} input image, predicted segments, fitted 3D body.}
\label{fig:pennaction_quantitative}
\end{figure*}

\subsubsection{Results on the H36M Dataset.}
H36M is a dataset recorded in a lab setting with
3D ground truth available through motion capture data.
We initialize with our model from the unsupervised training on the SURREAL dataset
and adapt it to H36M with minimal supervision (only 27 and 66 labeled
images) to limit the impact of the static background on our discriminator.
During training, we sample poses for domain \domB/ from the
same Gaussian mixture prior as
in~\cite{lassner2017unite}. Background images are taken from H36M
directly. The input image crops are generated following the
procedure in~\cite{omran2018neural}. We perform data augmentation by
rotating, mirroring and rescaling the images. For the few labeled
images, we extend the augmentation by pasting the segmented persons in
front of varying backgrounds.  We use subject 1 as a held-out test
subject and subjects 5, 6, 7 and 8 as training subjects.

\begin{table}
  \begin{center}
    \resizebox{0.48\textwidth}{!}{
      \begin{tabular}{|l|c|c|c|c|}
\cline{2-5}
\multicolumn{1}{c|}{} & IoU, 14 Seg. & IoU, 4 Seg. & IoU, 1 Seg. & tr-MPJPE \\
\hline
\cite{ramakrishna2012reconstructing}* & - & - & - & 157.3 \\
\hline 
SMPLify~\cite{Bogo:ECCV:2016}* & - & - & - & 82.3 \\
\hline 
NBF~\cite{omran2018neural}* & - & - & - & 59.9 \\
\hline
HMR~\cite{kanazawa2018end}* & - & - & - & 56.8 \\
\hline
\hline
Ours (27 labeled images) & 0.292 & 0.567 & 0.797 & 142.2 \\
\hline 
Ours (66 labeled images) & 0.326 & 0.568 & 0.800 & 140.1 \\
\hline 
      \end{tabular}
    }
  \end{center}
  \vspace*{-0.3cm}
  \caption{\textit{Results on the H36M dataset~\cite{h36m_pami}.} IoU for predicted body part segments (14 parts, 4 parts and fg/bg) and 3D error (mean per joint reprojection error after translation and rotation alignment) for varying amounts of supervision. *These results are computed on the full H36M test set and trained with supervision. We provide them to give an impression of the current supervised state of the art.}
  \label{tab:res_h36m}
\end{table}

Tab.~\ref{tab:res_h36m} shows quantitative results. We list results from supervised methods for comparison.  Notice that they are not strictly comparable, as the mean per joint reprojection error after translation and rotation alignment (tr-MPJPE) is calculated on H36M joints instead of SMPL joints and a different training / test split is used. For our method, left-right swaps are a major obstacle. However, the tr-MPJPE shows that the rough pose of the person is extracted faithfully. Under-represented poses like
sitting strongly impact the average error. In Fig.~\ref{fig:h36m}, we provide randomly picked qualitative results to give a direct impression of the predictions.

\subsubsection{Qualitative Results on In-The-Wild Images.}
H3.6M is a studio dataset, thus the variety of person appearances and
backgrounds is very limited. To show that our model can be used
on in-the-wild images, we provide additional qualitative results
for less constrained images. We fine-tune the unsupervised SURREAL
model on the Penn Action training set \cite{zhang2013actemes}. The
dataset provides bounding boxes; we use the regions around the people
to create background samples for domain \domB/. 3D pose and shape are
sampled from the same distribution as for SURREAL. Even though
Penn Action is a video dataset, we only train on individual,
unpaired images. Fig.~\ref{fig:pennaction_quantitative} shows example
results.
This highlights that our model does not over-fit to characteristics
of the synthetic SURREAL images, but does learn to represent
humans and recognize them in images.

\section{Conclusion}\label{sec:conclusion}

The 2D-3D generation-parsing cycle is ubiquitous in vision systems.
In this paper we have investigated a novel
model that aims to capture the entire cycle and use its structure
to enable unsupervised learning of the underlying relationships.
For this purpose, we have developed a chain of
cycles that link representations at different levels of
abstraction. We demonstrate the feasibility of \emph{representation cycling}, which we
think will be applicable to a variety of problems.
In this work we have focused on learning the extraction and generation of human bodies and,
to our knowledge, are the first to show entirely unsupervised results
on this problem. The complexity of the human body, self-occlusion,
and varying appearance all make this problem challenging, even with paired training data.
So why attack such a
hard problem? While getting supervised data for 3D human pose and
shape is still possible, it is much harder for other
object classes like animals.
Our model pushes how far we can go without paired data. It does not
yet match state-of-the-art results on human pose benchmarks---getting
there will likely require an engineering effort
that supervised models already have behind them.
Nonetheless, we see considerable potential in the longer term.
Approaches that learn on unlabeled data enable us to exploit almost
unlimited amounts of data and thus have the potential to yield more
robust models. Furthermore, an unsupervised approach can directly
be applied to new object categories, avoiding repeating
labeling efforts.

\vspace*{0.3cm}
\noindent\textbf{Aknowledgements.} \textit{This research was supported by the Max Planck ETH Center
for Learning Systems. MJB has received research gift funds
from Intel, NVIDIA, Adobe, Facebook, and Amazon. While
MJB is a part-time employee of Amazon, this research was
performed solely at MPI.}

{\small
\bibliographystyle{aaai}
\bibliography{AAAI-RueeggN.199}

\begin{thebibliography}{}

\bibitem[\protect\citeauthoryear{Almahairi \bgroup et al\mbox.\egroup
  }{2018}]{almahairi2018augmented}
Almahairi, A.; Rajeswar, S.; Sordoni, A.; Bachman, P.; and Courville, A.
\newblock 2018.
\newblock Augmented {CycleGAN}: Learning many-to-many mappings from unpaired
  data.
\newblock {\em arXiv preprint arXiv:1802.10151}.

\bibitem[\protect\citeauthoryear{Balakrishnan \bgroup et al\mbox.\egroup
  }{2018}]{balakrishnan2018synthesizing}
Balakrishnan, G.; Zhao, A.; Dalca, A.~V.; Durand, F.; and Guttag, J.
\newblock 2018.
\newblock Synthesizing images of humans in unseen poses.
\newblock {\em arXiv preprint arXiv:1804.07739}.

\bibitem[\protect\citeauthoryear{Bogo \bgroup et al\mbox.\egroup
  }{2016}]{Bogo:ECCV:2016}
Bogo, F.; Kanazawa, A.; Lassner, C.; Gehler, P.; Romero, J.; and Black, M.~J.
\newblock 2016.
\newblock Keep it {SMPL}: Automatic estimation of {3D} human pose and shape
  from a single image.
\newblock In {\em Proc. ECCV}.

\bibitem[\protect\citeauthoryear{Esser, Sutter, and
  Ommer}{2018}]{Esser_2018_CVPR}
Esser, P.; Sutter, E.; and Ommer, B.
\newblock 2018.
\newblock A variational u-net for conditional appearance and shape generation.
\newblock In {\em Proc. CVPR}.

\bibitem[\protect\citeauthoryear{He \bgroup et al\mbox.\egroup
  }{2016}]{he2016deep}
He, K.; Zhang, X.; Ren, S.; and Sun, J.
\newblock 2016.
\newblock Deep residual learning for image recognition.
\newblock In {\em Proc. CVPR}.

\bibitem[\protect\citeauthoryear{Hoffman \bgroup et al\mbox.\egroup
  }{2017}]{hoffman2017cycada}
Hoffman, J.; Tzeng, E.; Park, T.; Zhu, J.-Y.; Isola, P.; Saenko, K.; Efros,
  A.~A.; and Darrell, T.
\newblock 2017.
\newblock Cycada: Cycle-consistent adversarial domain adaptation.
\newblock {\em arXiv preprint arXiv:1711.03213}.

\bibitem[\protect\citeauthoryear{Hoshen and Wolf}{2018}]{hoshen2018nam}
Hoshen, Y., and Wolf, L.
\newblock 2018.
\newblock Nam - unsupervised cross-domain image mapping without cycles or
  {GANs}.

\bibitem[\protect\citeauthoryear{Huang \bgroup et al\mbox.\egroup
  }{2018}]{huang2018multimodal}
Huang, X.; Liu, M.-Y.; Belongie, S.; and Kautz, J.
\newblock 2018.
\newblock Multimodal unsupervised image-to-image translation.
\newblock {\em arXiv preprint arXiv:1804.04732}.

\bibitem[\protect\citeauthoryear{Ionescu \bgroup et al\mbox.\egroup
  }{2014}]{h36m_pami}
Ionescu, C.; Papava, D.; Olaru, V.; and Sminchisescu, C.
\newblock 2014.
\newblock Human3.6m: Large scale datasets and predictive methods for 3d human
  sensing in natural environments.
\newblock {\em PAMI}.

\bibitem[\protect\citeauthoryear{Jakab \bgroup et al\mbox.\egroup
  }{2018}]{jakab2018unsupervised}
Jakab, T.; Gupta, A.; Bilen, H.; and Vedaldi, A.
\newblock 2018.
\newblock Unsupervised learning of object landmarks through conditional image
  generation.
\newblock In {\em Adv. NeurIPS}.

\bibitem[\protect\citeauthoryear{Jakab \bgroup et al\mbox.\egroup
  }{2019}]{jakab2019learning}
Jakab, T.; Gupta, A.; Bilen, H.; and Vedaldi, A.
\newblock 2019.
\newblock Learning human pose from unaligned data through image translation.
\newblock In {\em Proc. CVPR}.

\bibitem[\protect\citeauthoryear{Kanazawa \bgroup et al\mbox.\egroup
  }{2018}]{kanazawa2018end}
Kanazawa, A.; Black, M.~J.; Jacobs, D.~W.; and Malik, J.
\newblock 2018.
\newblock End-to-end recovery of human shape and pose.
\newblock In {\em Proc. CVPR}.

\bibitem[\protect\citeauthoryear{Kingma and Welling}{2013}]{kingma2013auto}
Kingma, D.~P., and Welling, M.
\newblock 2013.
\newblock Auto-encoding variational bayes.
\newblock {\em arXiv preprint arXiv:1312.6114}.

\bibitem[\protect\citeauthoryear{Kudo \bgroup et al\mbox.\egroup
  }{2018}]{kudo2018unsupervised}
Kudo, Y.; Ogaki, K.; Matsui, Y.; and Odagiri, Y.
\newblock 2018.
\newblock Unsupervised adversarial learning of 3d human pose from 2d joint
  locations.
\newblock {\em arXiv preprint arXiv:1803.08244}.

\bibitem[\protect\citeauthoryear{Lassner \bgroup et al\mbox.\egroup
  }{2017}]{lassner2017unite}
Lassner, C.; Romero, J.; Kiefel, M.; Bogo, F.; Black, M.~J.; and Gehler, P.~V.
\newblock 2017.
\newblock Unite the people: Closing the loop between 3d and 2d human
  representations.
\newblock In {\em Proc. CVPR}.

\bibitem[\protect\citeauthoryear{Lassner, Pons-Moll, and
  Gehler}{2017}]{Lassner:GP:2017}
Lassner, C.; Pons-Moll, G.; and Gehler, P.~V.
\newblock 2017.
\newblock A generative model of people in clothing.
\newblock In {\em Proc. ICCV}.

\bibitem[\protect\citeauthoryear{Lee \bgroup et al\mbox.\egroup
  }{2018}]{lee2018diverse}
Lee, H.-Y.; Tseng, H.-Y.; Huang, J.-B.; Singh, M.~K.; and Yang, M.-H.
\newblock 2018.
\newblock Diverse image-to-image translation via disentangled representations.
\newblock {\em arXiv preprint arXiv:1808.00948}.

\bibitem[\protect\citeauthoryear{Loper \bgroup et al\mbox.\egroup
  }{2015}]{loper2015smpl}
Loper, M.; Mahmood, N.; Romero, J.; Pons-Moll, G.; and Black, M.~J.
\newblock 2015.
\newblock Smpl: A skinned multi-person linear model.
\newblock {\em ACM TOG}.

\bibitem[\protect\citeauthoryear{Lorenz \bgroup et al\mbox.\egroup
  }{2019}]{lorenz2019unsupervised}
Lorenz, D.; Bereska, L.; Milbich, T.; and Ommer, B.
\newblock 2019.
\newblock Unsupervised part-based disentangling of object shape and appearance.
\newblock {\em arXiv preprint arXiv:1903.06946}.

\bibitem[\protect\citeauthoryear{Ma \bgroup et al\mbox.\egroup
  }{2017}]{ma2017pose}
Ma, L.; Jia, X.; Sun, Q.; Schiele, B.; Tuytelaars, T.; and Van~Gool, L.
\newblock 2017.
\newblock Pose guided person image generation.
\newblock In {\em Adv. NeurIPS}.

\bibitem[\protect\citeauthoryear{Ma \bgroup et al\mbox.\egroup
  }{2018}]{ma2018disentangled}
Ma, L.; Sun, Q.; Georgoulis, S.; Van~Gool, L.; Schiele, B.; and Fritz, M.
\newblock 2018.
\newblock Disentangled person image generation.
\newblock In {\em Proc. CVPR}.

\bibitem[\protect\citeauthoryear{Madadi, Bertiche, and
  Escalera}{2018}]{madadi2018smplr}
Madadi, M.; Bertiche, H.; and Escalera, S.
\newblock 2018.
\newblock Smplr: Deep smpl reverse for 3d human pose and shape recovery.
\newblock {\em arXiv preprint arXiv:1812.10766}.

\bibitem[\protect\citeauthoryear{Martinez \bgroup et al\mbox.\egroup
  }{2017}]{martinez2017simple}
Martinez, J.; Hossain, R.; Romero, J.; and Little, J.~J.
\newblock 2017.
\newblock A simple yet effective baseline for 3d human pose estimation.
\newblock In {\em Proc. ICCV}.

\bibitem[\protect\citeauthoryear{Minderer \bgroup et al\mbox.\egroup
  }{2019}]{minderer2019unsupervised}
Minderer, M.; Sun, C.; Villegas, R.; Cole, F.; Murphy, K.; and Lee, H.
\newblock 2019.
\newblock Unsupervised learning of object structure and dynamics from videos.
\newblock {\em arXiv preprint arXiv:1906.07889}.

\bibitem[\protect\citeauthoryear{Mueller \bgroup et al\mbox.\egroup
  }{2018}]{Mueller_2018_CVPR}
Mueller, F.; Bernard, F.; Sotnychenko, O.; Mehta, D.; Sridhar, S.; Casas, D.;
  and Theobalt, C.
\newblock 2018.
\newblock Ganerated hands for real-time 3d hand tracking from monocular rgb.
\newblock In {\em Proc. CVPR}.

\bibitem[\protect\citeauthoryear{Omran \bgroup et al\mbox.\egroup
  }{2018}]{omran2018neural}
Omran, M.; Lassner, C.; Pons-Moll, G.; Gehler, P.~V.; and Schiele, B.
\newblock 2018.
\newblock Neural body fitting: Unifying deep learning and model-based human
  pose and shape estimation.
\newblock In {\em Proc. 3DV}.

\bibitem[\protect\citeauthoryear{Pavlakos \bgroup et al\mbox.\egroup
  }{2017}]{pavlakos2017volumetric}
Pavlakos, G.; Zhou, X.; Derpanis, K.~G.; and Daniilidis, K.
\newblock 2017.
\newblock Coarse-to-fine volumetric prediction for single-image 3{D} human
  pose.
\newblock In {\em Proc. CVPR}.

\bibitem[\protect\citeauthoryear{Pavlakos, Zhou, and
  Daniilidis}{2018}]{pavlakos2018ordinal}
Pavlakos, G.; Zhou, X.; and Daniilidis, K.
\newblock 2018.
\newblock Ordinal depth supervision for 3{D} human pose estimation.
\newblock In {\em Proc. CVPR}.

\bibitem[\protect\citeauthoryear{Popa, Zanfir, and
  Sminchisescu}{2017}]{dmhs_cvpr17}
Popa, A.; Zanfir, M.; and Sminchisescu, C.
\newblock 2017.
\newblock {Deep Multitask Architecture for Integrated 2D and 3D Human Sensing}.
\newblock In {\em Proc. CVPR}.

\bibitem[\protect\citeauthoryear{Ramakrishna, Kanade, and
  Sheikh}{2012}]{ramakrishna2012reconstructing}
Ramakrishna, V.; Kanade, T.; and Sheikh, Y.
\newblock 2012.
\newblock {Reconstructing 3d Human Pose from 2d Image Landmarks}.
\newblock {\em Proc. ECCV}.

\bibitem[\protect\citeauthoryear{Rhodin, Salzmann, and
  Fua}{2018}]{rhodin2018unsupervised}
Rhodin, H.; Salzmann, M.; and Fua, P.
\newblock 2018.
\newblock Unsupervised geometry-aware representation for 3d human pose
  estimation.
\newblock {\em arXiv preprint arXiv:1804.01110}.

\bibitem[\protect\citeauthoryear{Robinette, Daanen, and
  Paquet}{1999}]{robinette1999caesar}
Robinette, K.~M.; Daanen, H.; and Paquet, E.
\newblock 1999.
\newblock The caesar project: a 3-d surface anthropometry survey.
\newblock In {\em 2nd 3DIM}.

\bibitem[\protect\citeauthoryear{Sun \bgroup et al\mbox.\egroup
  }{2017}]{sun2017compositional}
Sun, X.; Shang, J.; Liang, S.; and Wei, Y.
\newblock 2017.
\newblock Compositional human pose regression.
\newblock In {\em Proc. ICCV}.

\bibitem[\protect\citeauthoryear{Tome, Russell, and
  Agapito}{2017}]{tome2017lifting}
Tome, D.; Russell, C.; and Agapito, L.
\newblock 2017.
\newblock Lifting from the deep: Convolutional 3d pose estimation from a single
  image.
\newblock {\em Proc. CVPR}.

\bibitem[\protect\citeauthoryear{Tulsiani, Efros, and
  Malik}{2018}]{mvcTulsiani18}
Tulsiani, S.; Efros, A.~A.; and Malik, J.
\newblock 2018.
\newblock Multi-view consistency as supervisory signal for learning shape and
  pose prediction.
\newblock In {\em Proc. CVPR}.

\bibitem[\protect\citeauthoryear{Tung \bgroup et al\mbox.\egroup
  }{2017}]{NIPS2017_7108}
Tung, H.-Y.; Tung, H.-W.; Yumer, E.; and Fragkiadaki, K.
\newblock 2017.
\newblock Self-supervised learning of motion capture.
\newblock In {\em Adv. NeurIPS}.

\bibitem[\protect\citeauthoryear{Varol \bgroup et al\mbox.\egroup
  }{2017}]{varol2017learning}
Varol, G.; Romero, J.; Martin, X.; Mahmood, N.; Black, M.~J.; Laptev, I.; and
  Schmid, C.
\newblock 2017.
\newblock Learning from synthetic humans.
\newblock In {\em Proc. CVPR}.

\bibitem[\protect\citeauthoryear{Varol \bgroup et al\mbox.\egroup
  }{2018}]{varol18_bodynet}
Varol, G.; Ceylan, D.; Russell, B.; Yang, J.; Yumer, E.; Laptev, I.; and
  Schmid, C.
\newblock 2018.
\newblock {BodyNet}: Volumetric inference of {3D} human body shapes.
\newblock In {\em Proc. ECCV}.

\bibitem[\protect\citeauthoryear{Zanfir \bgroup et al\mbox.\egroup
  }{2018}]{Zanfir_2018_CVPR}
Zanfir, M.; Popa, A.-I.; Zanfir, A.; and Sminchisescu, C.
\newblock 2018.
\newblock Human appearance transfer.
\newblock In {\em Proc. CVPR}.

\bibitem[\protect\citeauthoryear{Zhang, Zhu, and
  Derpanis}{2013}]{zhang2013actemes}
Zhang, W.; Zhu, M.; and Derpanis, K.~G.
\newblock 2013.
\newblock From actemes to action: A strongly-supervised representation for
  detailed action understanding.
\newblock In {\em Proc. CVPR}.

\bibitem[\protect\citeauthoryear{Zhu \bgroup et al\mbox.\egroup
  }{2017a}]{CycleGAN2017}
Zhu, J.-Y.; Park, T.; Isola, P.; and Efros, A.~A.
\newblock 2017a.
\newblock Unpaired image-to-image translation using cycle-consistent
  adversarial networks.
\newblock In {\em Proc. ICCV}.

\bibitem[\protect\citeauthoryear{Zhu \bgroup et al\mbox.\egroup
  }{2017b}]{zhu2017toward}
Zhu, J.-Y.; Zhang, R.; Pathak, D.; Darrell, T.; Efros, A.~A.; Wang, O.; and
  Shechtman, E.
\newblock 2017b.
\newblock Toward multimodal image-to-image translation.
\newblock In {\em Adv. NeurIPS}.

\bibitem[\protect\citeauthoryear{Zuffi \bgroup et al\mbox.\egroup
  }{2017}]{zuffi20173d}
Zuffi, S.; Kanazawa, A.; Jacobs, D.~W.; and Black, M.~J.
\newblock 2017.
\newblock 3d menagerie: Modeling the 3d shape and pose of animals.
\newblock In {\em Proc. CVPR}.

\end{thebibliography}
}

\end{document}